# Extraction of topological features from communication network topological patterns using self-organizing feature maps

W. Ali, R.J. Mondragón and F. Alavi

Different classes of communication network topologies and their representation in the form of adjacency matrix and its eigenvalues are presented. A self-organizing feature map neural network is used to map different classes of communication network topological patterns. The neural network simulation results are reported.

*Introduction:* The architecture of communication network mainly depends upon underlying topology of the network. Different topologies have different strategies of routing like internet topology has power-law structure so most of the data traffic attracts towards the well connected nodes. Also, some communication networks have highly dynamic topology like ad hoc networks, where topology changes at different time steps due to mobility of nodes. In this study, we considered three different types of communication network topologies as shown in Fig. 1; regular, random and scale-free network topologies.

In regular network topologies, all the nodes are uniformly connected with the same number of links. In particular, if the degree of each node is *r*, then the network is regular of degree *r*. In contrast, a random network is a graph in which the links are distributed randomly. The graph of a random network is assumed to have a complex topology, without any known organizing principles. The scale-free network topology has two major ingredients, growth and preferential attachment. It means that the network grows by the addition of new nodes and new nodes prefer to attach to the nodes that are already well connected. This leads to the power-law degree distribution of links per node. A scale-free network was used to model the internet topology [1].

*Recognition System:* The recognition system consists of a pre-processor and a neural net classifier as shown in Fig. 2. The input topological pattern is represented in the form of an adjacency matrix. The adjacency matrix of a communication network having *N* nodes contains *N x N* elements $A_{ij}$, whose value is $A_{ij} = A_{ji} = 1$ if nodes *i* and *j* are connected, and 0 otherwise. The adjacency matrix is submitted to the pre-processor. The job of the pre-processor is to create an output for an input topological pattern in such a way that the pre-processed output remains unchanged even if the orders of the labels of nodes are changed in the input topological pattern. The pre-processor consists of eigenvalues of an adjacency matrix block. The eigenvalues of an adjacency matrix block computes the spectral analysis of an adjacency matrix in the form of eigenvalues. Eigenvalues reveal certain network topological properties [2]. The Fig. 3 shows the adjacency matrices and their eigenvalues of two topological graphs. From Fig. 3, it can be seen that both Graph A and Graph B have same topology but their orders of the labels of nodes are different. This affects the adjacency matrices of the graphs since adjacency matrices of Graph A and Graph B do not match with each other. In order to solve this problem we compute the eigenvalues of an adjacency matrix because eigenvalues are the invariant to orders of the labels of nodes in the topological graph. We can see that both graph A and graph B have same eigenvalues irrespective of their adjacency matrices. So eigenvalues are used as an input to self-organizing feature map (SOFM) neural network classifier. Self-organizing feature map neural network learn to classify the input topological patterns and map the classes of communication network topologies during training process.

*Simulation Results:* In order to investigate the classification capability of self-organizing feature map neural network, it is trained on a training data set of nine hundred communication network topological patterns (300 patterns for each regular, random and scale-free network). Moreover, the regular network topological patterns represent the degree of node 3, 4 and 5. The size of a communication network is considered as a 56 node topology. Hence, each topological pattern represents 56x56 adjacency matrix and its 56 eigenvalues. The architecture of a self-organizing feature map neural network used during training consists of a single layer of 6 output neurons with hexagonal topology. The Fig. 4 shows the training vectors and initial network neurons' weight vectors before training around the perimeter of this figure. The initial network neurons' positions are all concentrated at the black spot at (-3.8117, 4.9388). The neural network is trained for 2000 epochs by applying the Kohonen learning rule [3]. During training neurons learn different classes of topological patterns and neurons' weight vectors move towards the various training vectors' groups as shown in Fig. 5. Simulation results show that self-organizing feature map neural network successfully classify the input topological patterns into four classes of communication network topologies. These are as follows;

**1)**  Regular network topologies with connectivity of node 3 and 4

**2)**  Regular network topologies with connectivity of each node 5

**3)**  Random network topologies

**4)**  Scale-free network topologies

Additional training may require for further classify the communication network topological patterns. Note that the training vectors and self-organizing feature map neurons' weight vectors contain 56 components and graphically, it is not possible to plot the vectors in 56 dimensions, so all the vectors are represented here in two dimensions.

*Conclusion:* It is possible to represent the real world communication network topology in the form of eigenvalues of an adjacency matrix. Eigenvalues reveal certain topological properties that can be directly related to the topology of communication networks. The self-organizing feature map neural network recognises different classes of communication network topologies by decreasing the distance between neurons' weight vectors and training patterns' groups.

**Authors' affiliations:**

W. Ali and RJ. Mondragón (Department of Electronic Engineering, Queen Mary, University of London, Mile End Road, London E1 4NS, United Kingdom)

E-mail: wajahat.ali@elec.qmul.ac.uk

F. Alavi (Department of Computer Science, Queen Mary, University of London, Mile End Road, London E1 4NS, United Kingdom)

**Figure Captions:**

Fig. 1 Interconnection network topologies

    (a) Regular network topology

    (b) Random network topology

    (c) Scale-free network topology

Fig. 2 Neural Network based recognition system

Fig. 3 Representation of network topology using adjacency matrix and its eigenvalues

Fig. 4 Position of neurons' weight vectors before training in the form of a black spot

Fig. 5 Topology of neurons and position of neurons' weight vectors after training

Figure 1

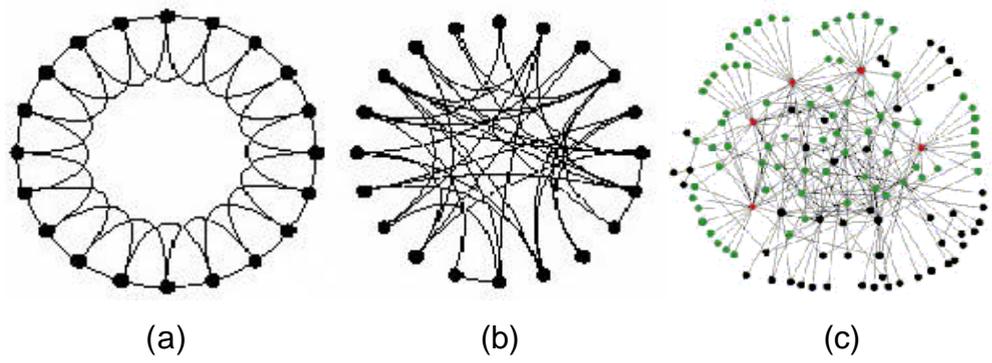

(a)　　　　　　　(b)　　　　　　　(c)

Figure 2

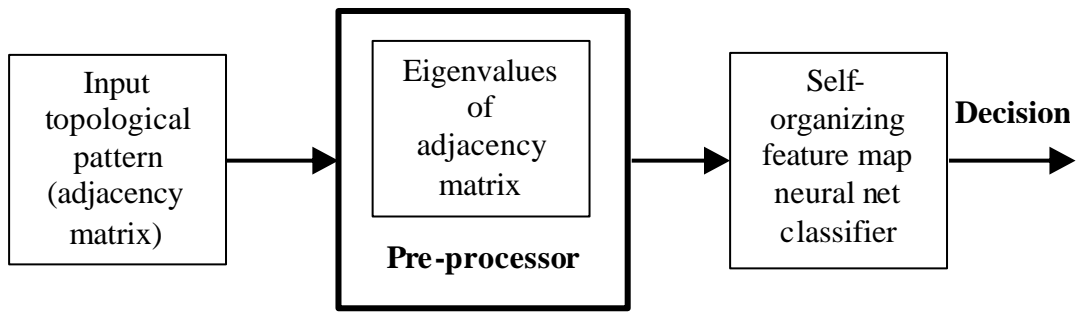

Figure 3

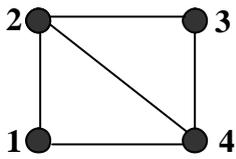 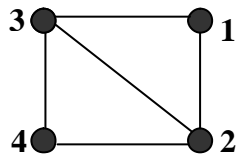

Graph A          Graph B

| 0 | 1 | 0 | 1 |
|---|---|---|---|
| 1 | 0 | 1 | 1 |
| 0 | 1 | 0 | 1 |
| 1 | 1 | 1 | 0 |

**Adjacency A**

| 0 | 1 | 1 | 0 |
|---|---|---|---|
| 1 | 0 | 1 | 1 |
| 1 | 1 | 0 | 1 |
| 0 | 1 | 1 | 0 |

**Adjacency B**

| $l_1$ | 1.5616 |
|---|---|
| $l_2$ | -1 |
| $l_3$ | 0 |
| $l_4$ | 2.5616 |

**Eigenvalues of Graph A**

| $l_1$ | 1.5616 |
|---|---|
| $l_2$ | -1 |
| $l_3$ | 0 |
| $l_4$ | 2.5616 |

**Eigenvalues of Graph B**

Figure 4

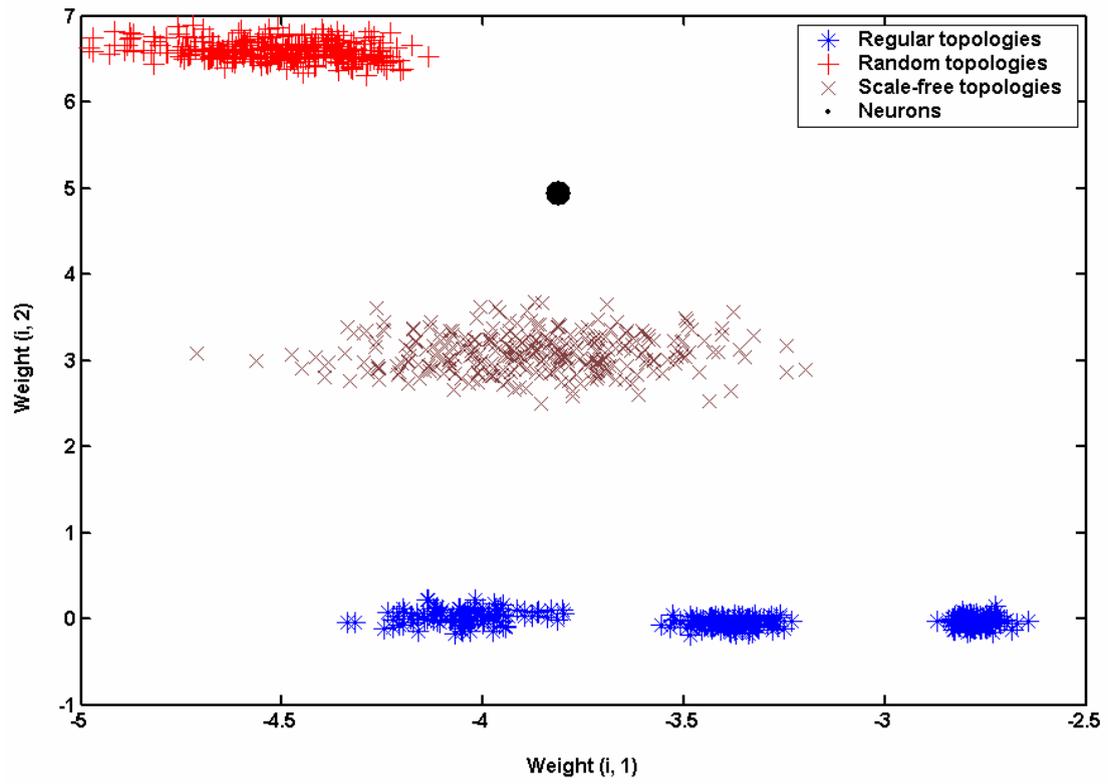

Figure 5

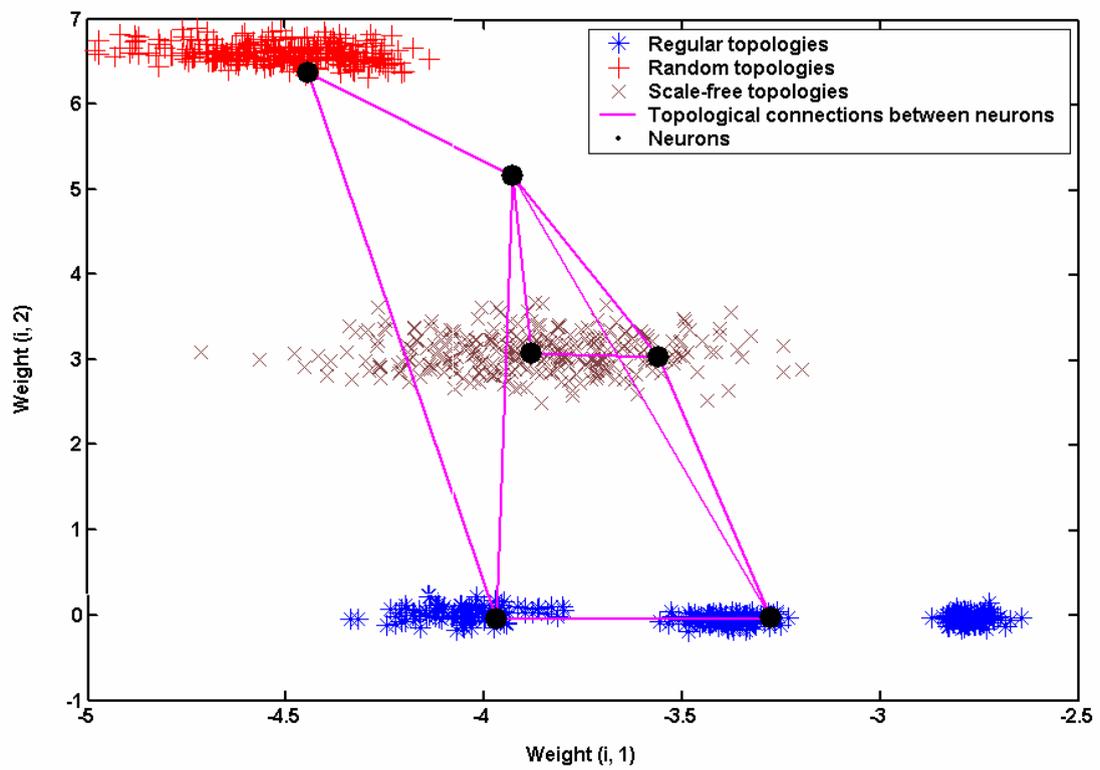